# Worth of knowledge in deep learning

Hao Xu[1], Yuntian Chen[2,*], and Dongxiao Zhang[2,3,*]

[1] BIC-ESAT, ERE, and SKLTCS, College of Engineering, Peking University; Beijing 100871, P. R. China
[2] Eastern Institute for Advanced Study, Eastern Institute of Technology; Ningbo, Zhejiang 315200, P. R. China
[3] Department of Mathematics and Theories, Peng Cheng Laboratory; Shenzhen 518000, Guangdong, P. R. China

* Corresponding authors. Email address: ychen@eias.ac.cn (Y. Chen); dzhang@eias.ac.cn (D. Zhang).

**Abstract:** Knowledge constitutes the accumulated understanding and experience that humans use to gain insight into the world. In deep learning, prior knowledge is essential for mitigating shortcomings of data-driven models, such as data dependence, generalization ability, and compliance with constraints. To enable efficient evaluation of the worth of knowledge, we present a framework inspired by interpretable machine learning. Through quantitative experiments, we assess the influence of data volume and estimation range on the worth of knowledge. Our findings elucidate the complex relationship between data and knowledge, including dependence, synergistic, and substitution effects. Our model-agnostic framework can be applied to a variety of common network architectures, providing a comprehensive understanding of the role of prior knowledge in deep learning models. It can also be used to improve the performance of informed machine learning, as well as distinguish improper prior knowledge.

**Main Text:**

The emergence of deep learning techniques has revolutionized the field of scientific research, resulting in a series of remarkable achievements (*1–3*). Deep learning excels at uncovering potential relationships in high-dimensional space from abundant available data. However, data-driven models still face certain challenges, such as data dependence (*4*), generalization ability (*5*), and compliance with constraints (*6*). In response to this, informed machine learning has become increasingly popular, enabling prior knowledge to be incorporated into the learning process (*7*, *8*). As illustrated in Fig. 1A, various types of knowledge can be integrated into a machine learning model, such as functional relations (*9*), logic rules (*10*), differential equations (*11*), invariance (*12*, *13*), and algebraic relations (*14*). For knowledge to be incorporated into a machine learning model, it needs to be formalized, meaning that it has to be structured in a manner that can be expressed mathematically. In this case, the formalized knowledge that can be integrated into a machine learning model is referred to as *rules*. Informed machine learning has been deployed in a variety of problem domains, such as the solution of partial differential equations (PDEs) (*15*, *16*), quantification of fluid flow (*4*), time series prediction (*17*), and robot control (*18*). Depending on how the importance of the rules is perceived, two main approaches in the field of informed machine learning are soft constraint (*19*) and hard constraint (*20*, *21*). Despite its promise, the worth of knowledge is currently only vaguely understood, which limits our ability to comprehend the relationship between data and knowledge. Fig. 1B provides a clear example of the divergent effects of data and rules in the context of interpolation and extrapolation. The data-driven model performs well in interpolation tasks, but poorly in extrapolation tasks. In contrast, the rule-driven model is able to handle the extrapolation task, while also maintaining moderate accuracy in both the extrapolation and interpolation tasks. This demonstrates the power of informed machine learning in leveraging both observed data and prior rules to achieve a universal prediction ability.

To elucidate the importance of data and rules, it is useful to consider the perspective of learning high-dimensional sample distributions (as illustrated in Fig. 1C). Conventional data-driven machine learning models attempt to learn the underlying sample distribution solely from observed data. However, this approach results in limited generalization ability, leading to poor performance when dealing with out-of-distribution (OOD) data (i.e., the extrapolation problem or OOD generalization). One potential solution to this problem is the incorporation of prior knowledge, which provides a restricted high-dimensional space of possible distributions that conform to the rules. While rules can reduce the optimization space of distributions, they cannot directly yield an accurate distribution unless a unique solution can be obtained from the prior rules. In this way, informed machine learning uses training data to locate the correct sample distribution within the restricted high-dimensional space created by the rules. The combination of data and rules can theoretically result in high efficiency and good inference ability for both interpolation and extrapolation tasks. However, incorporating multiple prior rules can lead to a high risk of model collapse due to intricate internal interactions, making convergence of the training process more difficult. To address this, it is necessary to measure the importance of each integrated rule to guide model construction and maximize the value of knowledge. In this work, our primary focus is on quantitatively measuring the worth of knowledge to uncover the underlying principles of data and rules. Our contributions can be summarized as addressing the following three main questions:

1. **How to evaluate the worth of knowledge?** Drawing inspiration from the Shapley value (*22*, *23*), we propose a framework for quantitatively measuring the effect of prior rules in informed machine learning. We introduce the concept of rule importance (abbreviated as *RI*) to effectively address the issue of assigning the contribution of integrated rules in informed deep learning. A detailed description of the definition and derivation of *RI* can be found in the Materials and Methods section.

2. **What is the relationship between data and rules?** Through quantitative experiments, we have evaluated the influence of data volume and estimation range on the worth of knowledge and uncovered the complex relationship between data and knowledge, comprising dependence, synergism, and substitution effects.

3. **How to make prior rules work better?** Our measurement of rule importance can be used to facilitate adjustment of the regularization parameter in informed machine learning to prevent non-convergence during the training process and maximize the value of knowledge. It can also be used to identify improper prior rules.

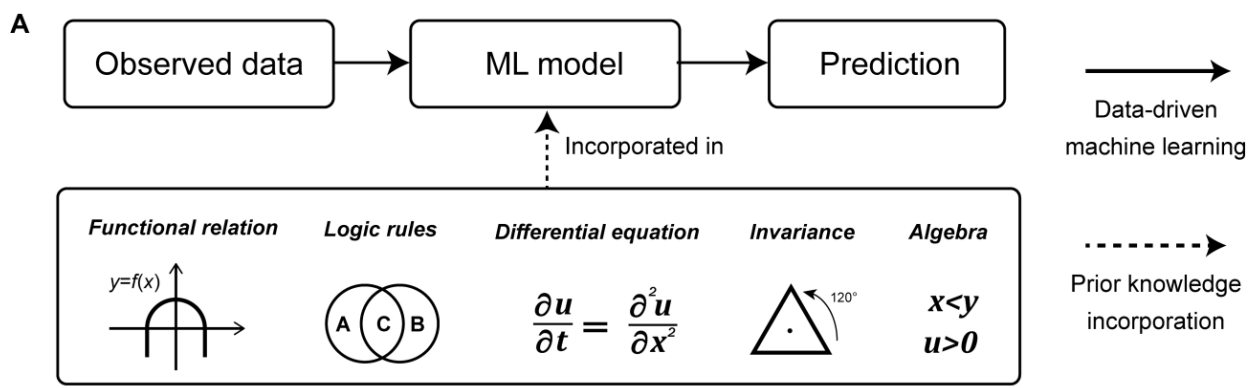

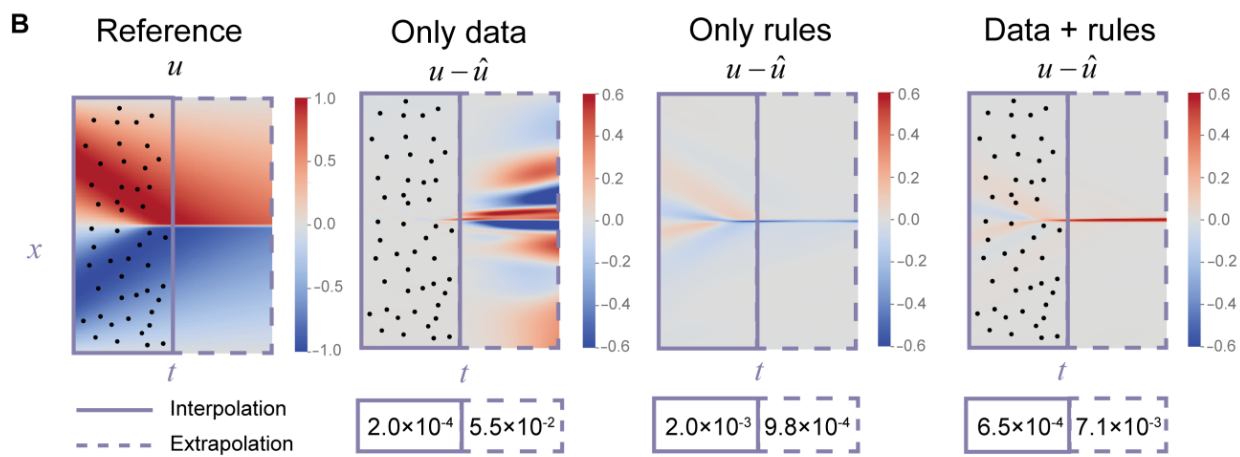

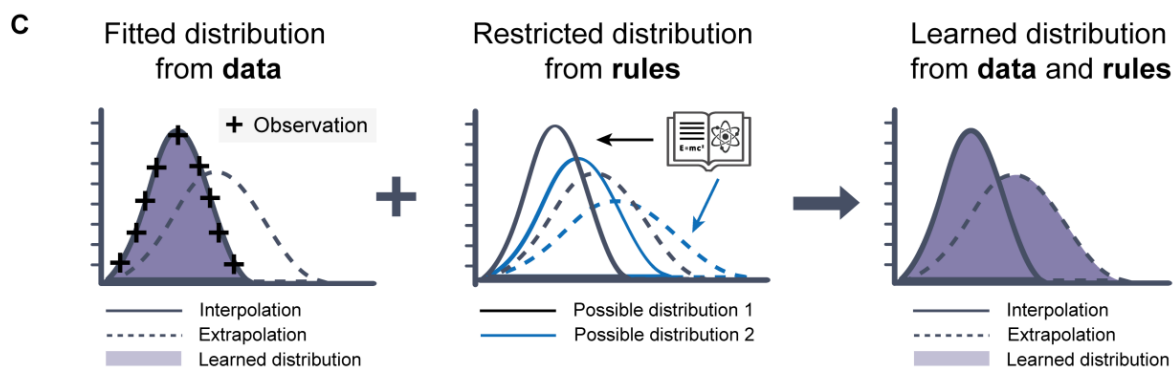

**Fig. 1. Overview of informed machine learning. (A)** The information flow in conventional data-driven machine learning and informed machine learning. **(B)** An example of Burgers' shock equation for informed machine learning. The black dots refer to the training data. $u$ and $\hat{u}$ are references and predictions, respectively. The mean squared error of interpolation and extrapolation is displayed in solid and dotted purple lines, respectively. **(C)** The explanation for the function of data and rules in informed machine learning from the aspect of the sample distribution. Here, rules refer to the formalized knowledge that can be incorporated into the machine learning model.

**Inherent principles behind data and rules**

In this section, we conduct a series of systematic experiments to elucidate the fundamental principles underlying data and rules. At this stage, four canonical physical processes that can be described by explicit governing equations are taken as representative examples, which are illustrated in Fig. 2A. The involved rules cover the governing partial differential equations (PDEs), boundary conditions (BCs), and initial conditions (ICs). The rules are incorporated into the model through the loss function (*7*), and the experimental settings are detailed in Supplementary Information S1. In the following part, we will provide a summary of qualitative and consistent insights that have been further explicated from the perspective of the sample distribution. These insights will also be validated via numerical experiments.

*In-distribution prediction: larger data volume, lower rule importance*

First, we examined the in-distribution prediction scenario, wherein the distribution of test data and training data is either similar or identical. Constructing the surrogate model is a typical in-distribution task since it aims to recover attributes in the whole domain through a few scattered observation data. Here, different volumes of data, including 0 (no data), 10, $10^2$, $10^3$, and $10^4$, are randomly sampled as the training dataset, and the test data volume is 10,000. For the four canonical physical processes, the importance of each integrated rule is calculated and illustrated in Fig. 2B and Fig. S7A.

Through the quantification of rule importance, the efficacy of different types of rules and their sensitivity to the data volume are reflected explicitly. Notably, we observe a diminishing impact of rule importance with increasing data volume, with distinct decay patterns depending on the type of rule (as illustrated in Fig. 2B). Our findings have uncovered counter-intuitive results in which, when the volume of data exceeds a certain threshold, the effect of PDEs may become negative. From the perspective of high-dimensional sample distribution, these findings can be elucidated since the model essentially interpolates on the learned sample distribution from training data for in-distribution prediction. In this scenario, a larger volume of data enables the model to learn a more accurate distribution without the need to integrate rules to limit the range of alternative distributions.

*Out-distribution prediction: larger data volume, higher global rule importance, lower local rule importance*

In real-world scenarios, physical processes often exhibit evolving distributions of quantities, which poses a challenge in predicting future quantities. Therefore, we investigate the influence of data volume on out-distribution prediction, which involves inferring a different sample distribution, such as future prediction. The experimental settings are provided in Supplementary Information S1.

In out-distribution prediction, we find that the influence of data volume varies with rule type. Rules that globally restrict the entire domain (e.g., PDE rules) are termed *global rules*, while rules that locally restrict the observable area (e.g., IC rules) are termed *local rules*. It is found that the importance of global rules increases as the data volume rises (Fig. 2C). Conversely, the importance of local rules diminishes with a larger data volume (Fig. 2C). This cognition is distinctive, but reasonable, since the unobservable domain outside of the distribution is difficult to learn with observable data. Under this circumstance, more observed data do not necessarily improve the predictive ability of the model, but instead increase the risk of overfitting. Therefore, global rules will play a more critical role in instructing the model by restricting possible distributions globally. In contrast, the function of local rules coincides with observed data, which accounts for the declining *RI* with data volume.

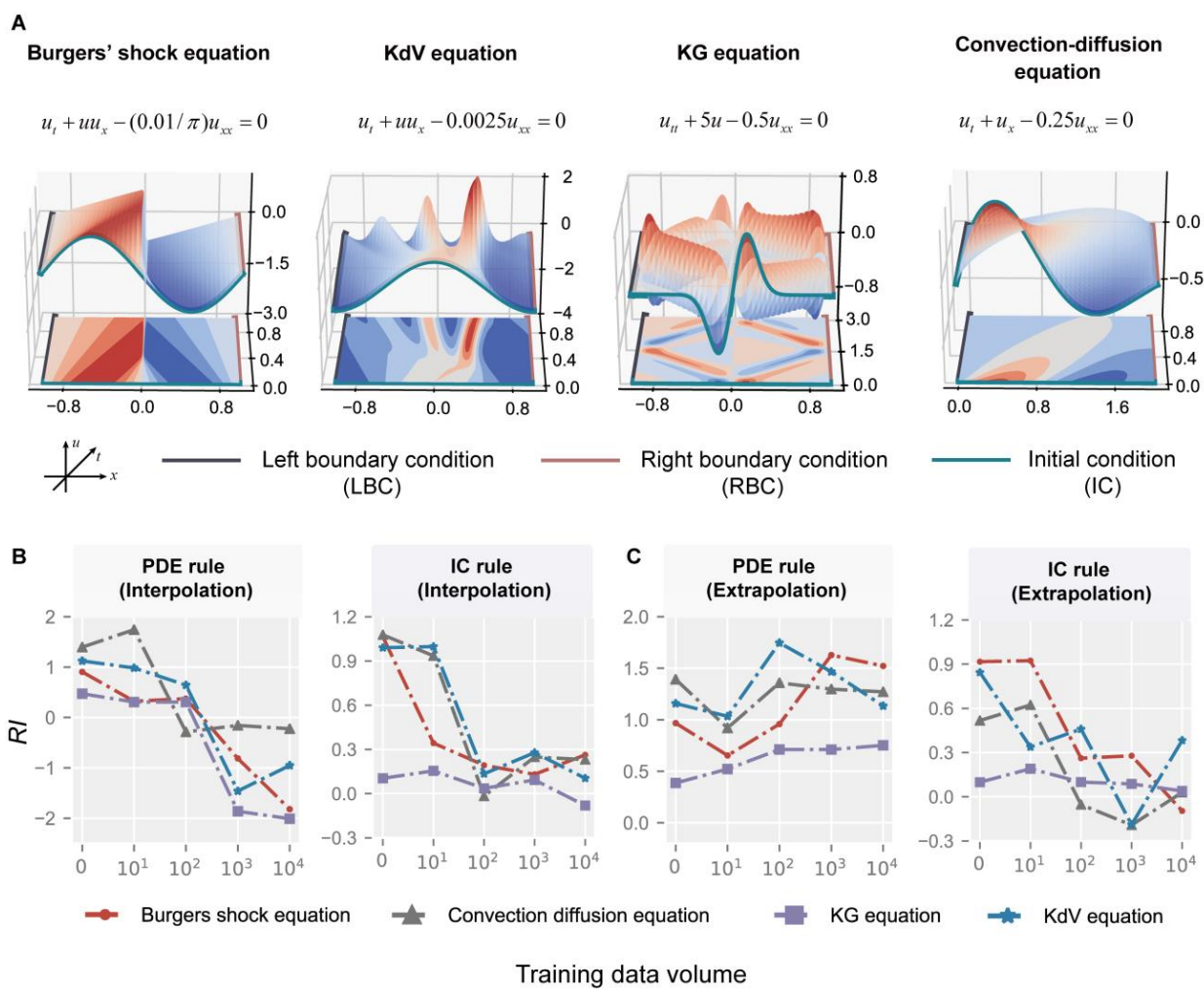

**Fig. 2. Numerical experiments to reveal the relationship between data and rules in the context of in-distribution (interpolation) and out-distribution (extrapolation) scenarios.** **(A)** The information of the four canonical physical processes that are guided by governing equations. The involved rules include PDEs, initial conditions, and boundary conditions. **(B)** The importance of PDE rules (global rules) and IC rules (local rules) varies with the data volume in the case of in-distribution in the four physical processes. **(C)** The importance of PDE rules (global rules) and IC

rules (local rules) varies with the data volume in the case of out-distribution in the four physical processes. The rule importance (*RI*) values are calculated using the framework established in this study.

**The interactions between rules**

In this section, we aim to explore the interactions among multiple rules through complex scenarios, such as the solution of multivariable equations and two-dimensional PDE which involves more rules. For solving multivariable equations, the following example is provided:

$$\begin{cases} c = |\sin(a) - \cos(b)| & (\text{rule 1}) \\ d = \log((a-b)^2 + 1) & (\text{rule 2}) \\ e = 0.5(1 + c^2) & (\text{rule 3}) \\ f = \exp(-e) & (\text{rule 4}) \end{cases}, \quad (1)$$

where the independent variables are $a \in [0, \pi]$ and $b \in [-\pi, 0]$; and the dependent variables are $c$, $d$, $e$, and $f$. Another implicit rule, $e, f > 0$ (rule 5) is also considered. The form of the complex two-dimensional PDE is written as:

$$\frac{\partial^2 u}{\partial x^2} - \frac{\partial^4 u}{\partial y^4} = (2 - x^2)e^{-y}, \quad (2)$$

where the definition domains are $x \in [0,1]$ and $y \in [0,1]$. There are six conditions for solving this PDE (termed as rule 1), namely:

$$\begin{array}{ll} u_{yy}(x,0) = x^2 \quad (\text{rule 2}) & u_{yy}(x,1) = x^2/e \quad (\text{rule 3}) \\ u(x,0) = x^2 \quad (\text{rule 4}) & u(x,1) = x^2/e \quad (\text{rule 5}) \\ u(0,y) = 0 \quad (\text{rule 6}) & u(1,y) = e^{-y} \quad (\text{rule 7}) \end{array}. \quad (3)$$

From our evaluation framework, several inherent interactions between rules are revealed explicitly, i.e., dependence, synergism, and substitution effects.

- Dependence

This work reveals the existence of extensive dependencies among rules and variables. Two types of dependence are identified: inner dependence between rules, and outer dependence between rules and dependent variables.

The inner dependence can be explicitly revealed by comparing the *full importance* (*FI*), which is defined as the importance of a given rule when all of the other rules exist, with the calculated rule importance *RI*. In the task of solving two-dimensional PDEs, as illustrated in Fig. 3A, the *RI* of most rules is lower than *FI*, especially the PDE rule (rule 1), indicating that most rules have dependence and need other relying rules to be effective. Considering that the proposed *RI* essentially measures the marginal contribution of the rule in all scenarios, the low *RI* of the PDE rule means that the incorporation of the PDE rule is ineffective in most scenarios since it depends highly on the other rules (i.e., high dependence), while the high *FI* indicates that it plays a significant role with complete relying rules. Moreover, it is observed that the inclusion of observation data affects inner dependence (Fig. 3A).

The outer dependence can be clearly elucidated through the framework proposed in this work,

which involves evaluating the importance of each involved rule for every dependent variable. As illustrated in Fig. 3B, the outer dependence in the task of solving multivariable equations is transparent, where a higher $RI$ corresponds to a larger dependence. For example, variable $d$ depends heavily on rule 2, while variable $f$ is slightly dependent on rules 1, 3, and 4. This result is consistent with the intuitive analysis of the rules, demonstrating the reliability of the proposed method, and indicating that the quantitative measurement of rule importance facilitates an explicit investigation of outer dependence.

- Synergism

Synergism is a specific form of interaction in which multiple rules work together to produce an effect that is greater than the sum of their individual effects. In our study, we observed evidence of synergism through the importance of different numbers of other relying rules, which is defined as:

$$RI_i^r(v) = \frac{1}{N_{S_i^r}} \sum_{s \in S_i^r} \log_{10}\left(\frac{MSE(s \setminus \{i\})}{MSE(s)}\right), \qquad (4)$$

where $RI_i^r$ is the importance of rule $i$ with the number of other relying rules $r$; and $S_i^r$ refers to the set of rules coalition containing the rule $i$ and other relying rules $r$. As illustrated in Fig. 3C and Fig. 3D, synergism is discovered in both cases. Specifically, in the task of solving multivariable equations, $RI^r$ shows an evident increase for rules 1, 2, 3, and 4, which indicates that a synergism effect exists in these rules. Conversely, rule 5 does not participate in the synergism. Similarly, in the task of solving two-dimensional equations, synergism is observed in the PDE rule. However, this synergy effect is only apparent when sufficient dependency rules exist (over five relying rules), and is not affected by the inclusion of observation data. The $RI^r$ for other rules is displayed in Fig. S9 and S10.

- Substitution

The substitution effect refers to the phenomenon in which the function of one rule can be substituted by either the data or other rules. This effect is illustrated in Fig. 3C, where the $RI^r$ of rule 5 is subtle overall and even decreases with a larger number of relying rules. We have discovered that rules regarding domain are more susceptible to being substituted when there are adequate data or sufficient relying rules. The substitution effect can arise when there are sufficient data, redundancy among rules, or when specific rules only apply to certain conditions or domains. Overall, apprehending the substitution effect can assist to simplify the model and improve its efficiency by reducing redundant or interchangeable rules.

**Practical applications**

In current machine learning research, it is widely acknowledged that incorporating prior knowledge can enhance the predictive ability of an informed machine learning model (*7*). However, the selection of appropriate prior knowledge and the quantification of its worth remain challenging, hindering the practical application of such techniques, particularly in scenarios with multiple rules. Our proposed framework manages to solve the above-mentioned problem and offers a solution at the pragmatic level. In this work, we present two examples of how our framework can be used to adjust the regularization parameter and distinguish improper prior rules.

One of the key challenges in informed machine learning, particularly in physics-informed neural networks (PINNs), is the difficulty of adjusting the weights of rules during the training process (*24*, *25*). Our framework provides a simple, yet efficient, strategy to address this problem, whereby we increase the weights of rules with positive importance and decrease the weights of rules with negative importance. The detailed algorithm for this process is provided in Fig. S11. We adopt the example of solving multivariable equations to demonstrate the effectiveness of our technique, and the optimized results are presented in Fig. 4A and 4B. The prediction error shows a significant decrease during the optimization (Fig. 4A), indicating its efficacy. Meanwhile, the importance of most rules increases after the optimization, which indicates that the adjustment assists to enhance the value of knowledge (Fig. 4B). The comparison with other optimization methods is provided in Table S1.

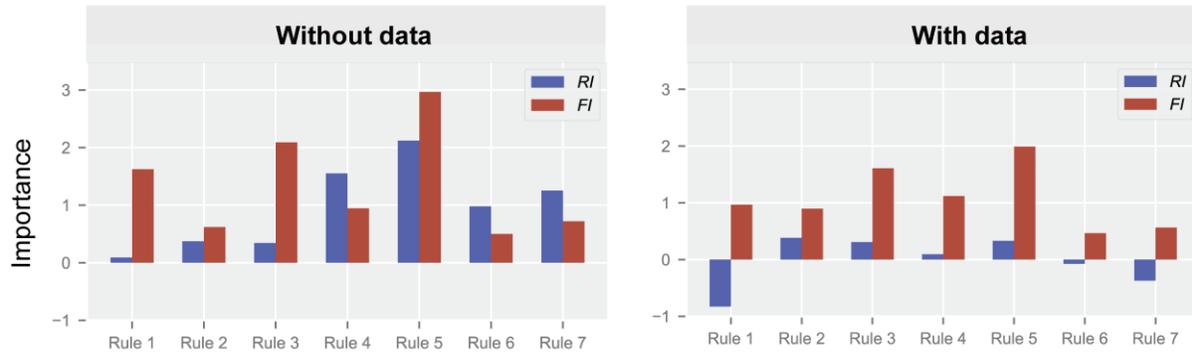

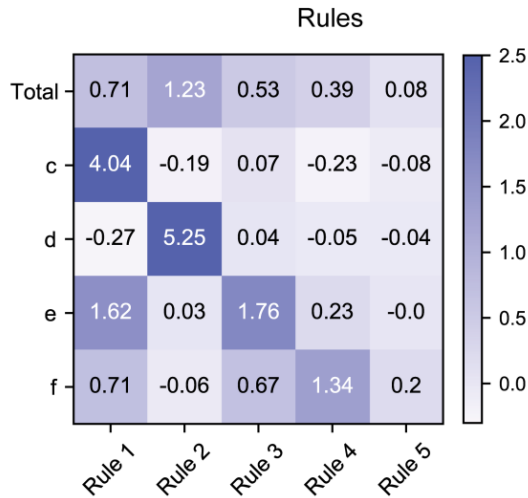

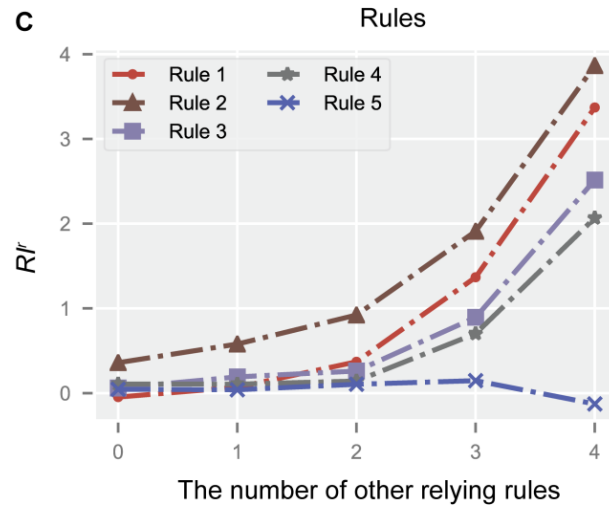

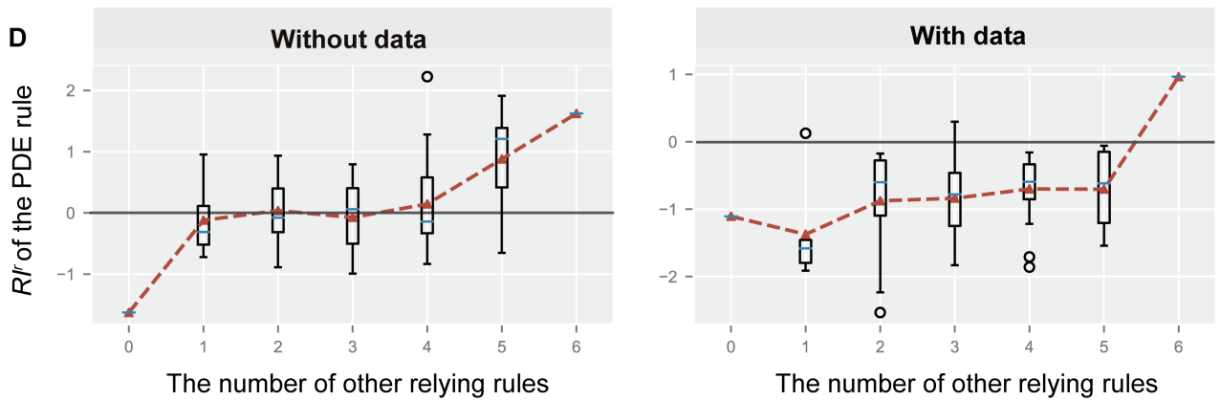

**Fig. 3. Numerical experiments to explore the interactions among multiple rules. (A)** The rule importance (*RI*) and full importance (*FI*) of each involved rule when solving a two-dimensional PDE with and without data. **(B)** The *RI* of the involved rules is calculated by each dependent variable when solving multivariable equations without data. **(C)** The importance of the involved rules under different numbers of relying rules when solving multivariable equations without data. **(D)** The importance of the PDE rule under different numbers of relying rules when solving two-dimensional PDE with and without data. The boxplot refers to all calculated importance with a given number of relying rules. The blue lines in the boxplot refer to the median.

This framework can also assist in identifying improper or wrong rules that may negatively impact the model. In real-world applications, it is often challenging to ensure that all rules are appropriate for the given problem, which can interfere with model training. A numerical example is provided in Fig. 4C, in which each rule is mistaken by a wrong rule with an imperceptible error in turn. The Burgers' shock equation is utilized here as an example, and the change in the *RI* of each rule in the presence of wrong rules is also displayed in Fig. 4C. Several interesting findings can be obtained from the results. First, the importance of the wrong rules presents a fully negative effect, leading to a significant decline in importance. This striking negative importance highlights the wrong rule, which provides an explicit way to identify the improper rules. Second, the wrong rules will affect the importance of the other rules. For example, the importance of the PDE rule (rule 1) decreases with the existence of other improper rules. In contrast, the importance of the initial rule (rule 2) increases when the boundary conditions are wrong, but decreases with a wrong PDE rule. This phenomenon implies the existence of different competitive and promotional relationships between rules.

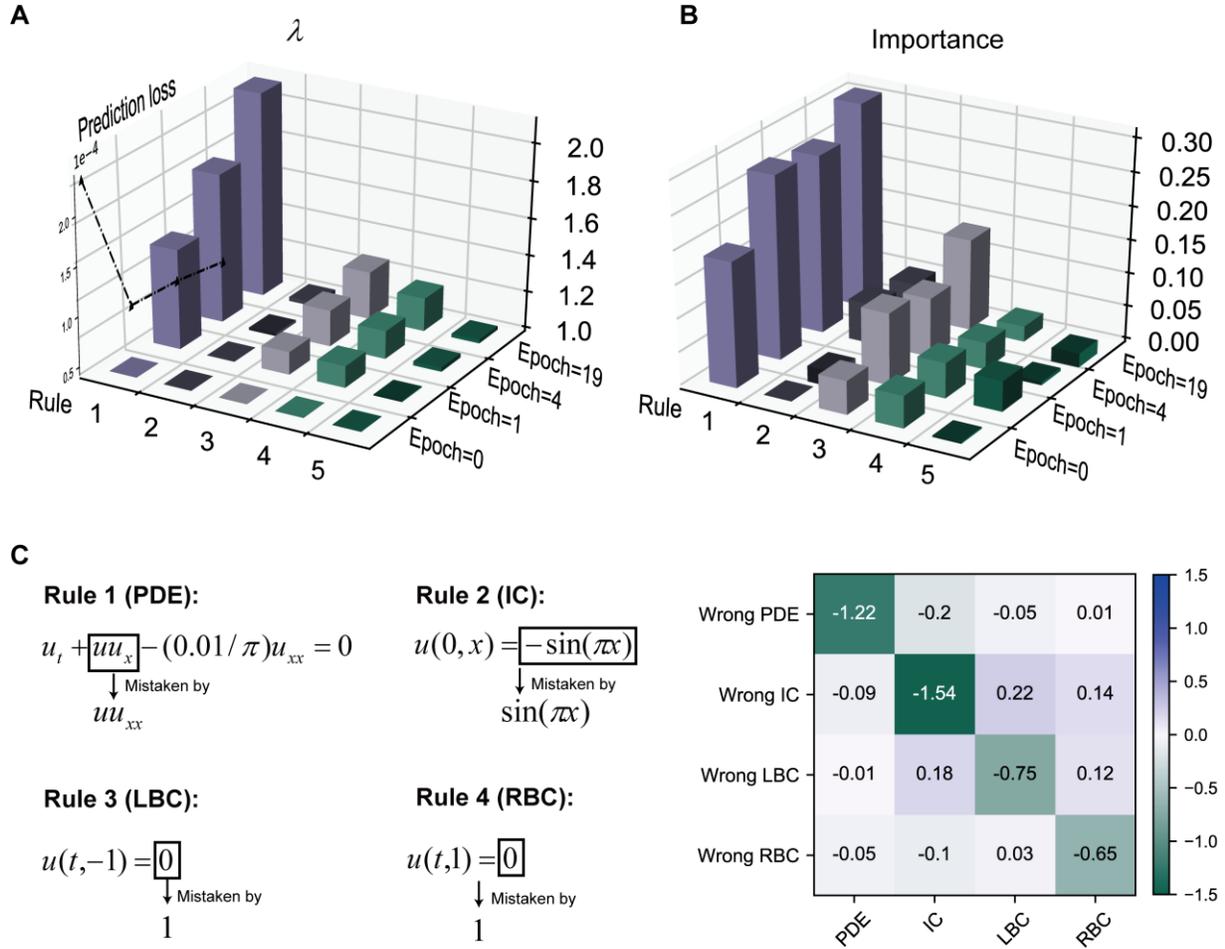

**Fig. 4. Practical application of rule importance. (A)** The change in the weights $\lambda_i$ of rules under different iterations. The left side is the prediction loss measured by mean squared error. The example of solving multivariable equations is utilized here. **(B)** The change of rule importance during the optimization process. **(C)** The correct rules and the corresponding mistaken part (left), and the change in the importance of each rule with wrong rules (right). Here, the situation in which only one rule error is considered.

## Conclusion

In this study, we have introduced an interpretable framework for evaluating the worth of knowledge in deep learning. The proposed measurements of *FI* and *RI* based on Shapley value provide insights into the relationship between data and rules. We have also investigated the dependence, synergies, and substitution of rules, which can assist to identify important knowledge components. Additionally, the importance of rules can be utilized in practical applications, such as improving the performance of informed machine learning and identifying improper rules, which in turn can enhance the worth of knowledge. We have also confirmed the significance of this framework through a practical engineering problem regarding 2D heterogeneous subsurface flow (Supplementary Information S2.3). Moreover, we evaluated the effectiveness of the framework in the presence of noisy data, which discovers that prior rules tend to have a higher importance in the presence of moderate data noise (Supplementary Information S2.4). Meanwhile, it is revealed that the choice of collocation points, including the location and volumes, has an impact on the model training (Supplementary Information S2.1).

In the future, we will further speed-up the calculation process of rule importance through the Monte Carlo-type method since it involves training multiple models in parallel with different rule coalitions, which may take a relatively long time (0.5~2 h). Additionally, we aim to extend the proposed framework to measure the importance of rules related to mathematics and physics, such as invariance and logic rules, and investigate their functions in deep learning. Overall, we believe that evaluating the worth of knowledge is critical not only for interpretability, but also for improving the security and reliability of informed machine learning models. This approach is especially relevant for future research on large-scale models, in which model security is a growing concern.

**Acknowledgments:**

**Funding:** This work was supported and partially funded by:

The National Center for Applied Mathematics Shenzhen (NCAMS)

The Shenzhen Key Laboratory of Natural Gas Hydrates (Grant No. ZDSYS20200421111201738)

The SUSTech – Qingdao New Energy Technology Research Institute

The National Natural Science Foundation of China (Grant No. 62106116)

**Author contributions:** H. X., Y. C., and D. Z. conceived the project, designed and performed research, and wrote the paper; H. X. implemented workflow, created code, visualized results, and analyzed and curated data.

**Competing interests:** Authors declare that they have no competing interests.

**Data and materials availability:** The dataset utilized in this study has been deposited in the GitHub repository, https://github.com/woshixuhao/Worth_of_Knowledge/tree/main/Data. All original code has been deposited in the GitHub repository, https://github.com/woshixuhao/Worth_of_Knowledge/tree/main/Code.

## Supplementary Materials

Materials and Methods

Supplementary Text

Figs. S1 to S11

Table S1

References (*26–30*)

# Supplementary Materials for

## Worth of knowledge in deep learning

Hao Xu[1], Yuntian Chen[2,*], and Dongxiao Zhang[2,3,*]

* Corresponding authors: ychen@eias.ac.cn (Y. Chen); dzhang@eias.ac.cn (D. Zhang).

**The PDF file includes:**

    Materials and Methods
    Supplementary Text
    Figs. S1 to S11
    Table S1
    References *(26–30)*

**Materials and Methods**

**Shapley value**

Shapley value was first proposed by Shapley in 1951 and evolved into a commonly-used indicator to fairly and quantitatively evaluate the marginal contribution of participants (*22*). This indicator originated from cooperative games and has been used in a wide range of fields. In machine learning, the Shapley value is adapted to choose potential characteristics, rank the importance of training data, and improve the interpretability of the model (*26*).

First, for $N$ different participants, the collection of participants $I$ is denoted as $I = \{1,2,3,\ldots,N\}$. For each subset $s$ in $I$, the profit of the alliance $s$ is defined as $v(s)$. Shapley value is established on the premise that cooperation is at least more beneficial than the participants alone or in small groups (i.e., coalitional game). The opposite is a non-coalitional game, in which one participant's profit will inevitably lead to the other participant's deficit. The target of the Shapley value is to provide a criterion to determine the contribution of each participant $\varphi_i(v)$ in the cooperation in a fair manner. Shapley proves that under the premise that the total profit of all participants equals the profit of cooperation (i.e., $\sum_{i \in s} \varphi_i(v) = v(I)$), the unique solution is written as:

$$\varphi_i = \sum_{s \in S_i} w(|s|)[v(s) - v(s \setminus \{i\})], . \tag{S.1}$$

where $\varphi_i$ is the Shapley value of the participant $i$; $S_i$ is the set of all possible combinations including the member $i$; $s\setminus\{i\}$ is the new set produced by removing the member $i$ from $s$; $v$ is the contribution; and $w(|s|)$ is the weight. The weight can be calculated by the following formula:

$$w(|s|) = \frac{(|s|-1)!(n-|s|)!}{n!} . \tag{S.2}$$

It is worth noting that Eq. (S.2) is obtained according to the principle of permutation, which means that the order of arrival is considered. Here, we define the marginal contribution of participant $i$ when incorporating into $S$ as:

$$\delta_i(s) = v(s) - v(s \setminus \{i\}), s \in S_i . \tag{S.3}$$

Then, Eq. (S.1) can be converted into:

$$\varphi_i = \sum_{s \in S_i} \frac{(|s|-1)!(n-|s|)!}{n!} \delta_i(s) . \tag{S.4}$$

Therefore, the process of calculating the Shapley value can be roughly summarized as calculating the average value of marginal contribution in nodes of $n!$ situations in all sequence permutations.

**The theoretical basis for rule importance**

For the simplest case, in which a single rule is integrated into the model, the profit of integrating this rule can be easily measured by the improvement of the network's predictive accuracy. However, informed machine learning usually involves multiple prior rules, which will incur complex interactions between rules. Under this circumstance, the measurement of rule importance is essentially the assignment of profits in a coalitional game, in which prior rules work together to influence the model's predictive ability. Inspired by the idea of Shapley value, which is initially proposed to solve the problem of determining individual contribution in the coalitional game (*22, 23*) (Supplementary Information S1.1), a framework for measuring rule importance in the informed machine learning model is established. For the calculation of rule importance, several descriptions are put forward beforehand to describe the process explicitly, which can be summarized as follows:

1. The basic predictive performance of the model without any rules is seen as the baseline

In the Shapley value, it is assumed that the total profit of the alliance is zero if no one contributes. In this work, rules are incorporated into the model to improve predictive accuracy. Therefore, the basic predictive performance of the model without any rules is seen as the baseline. In other words, the predictive accuracy of the model trained with only data (or even no data) is employed for comparison to measure the effect of incorporating rules.

2. The profit of the rule coalition is defined as the improvement of the model's predictive accuracy

In this work, the profit of the rule coalition is embodied by the improvement of the model's predictive accuracy. The definition means that negative contributions may exist due to the inclusion of improper rules or their coalition, which may actually lead to a decrease in overall predictive accuracy. Therefore, the calculated rule importance can be either positive or negative, as rules may improve or decrease the predictive accuracy of the model. It is important to note that this problem is inherently complex, as it involves a coalitional game, in which the alliance of rules may not always enhance the profit. As a result, there may be complex interactions between rules involving both cooperation and competition, which is carefully considered in our analysis.

3. The importance of rules is assigned by their respective marginal contributions

Inspired by the Shapley value, the proposed rule importance is also assigned by respective marginal contributions. This means that the measurement of rule *i* is essentially the average value of the difference in the model's predictive accuracy between all rule coalitions including and excluding rule *i*.

**Derivation of rule importance**

Inspired by the basic concept of Shapley value, the rule importance is proposed in this work to measure the worth of knowledge. In essence, the importance of rules is assigned by their respective marginal contribution where the profit of the rule coalition is defined as the improvement of the model's predictive accuracy. On the basis of the basic principles described above, the formula to calculate rule importance can be derived. To calculate the rule importance, the predictive performance of the model with all relevant rule coalitions is required, which is measured by the mean squared error (MSE). Similar to Shapley value, the marginal contribution of rule *i* when incorporating into *S* is:

$$\delta_i^{RI}(s) = -[\log_{10}(MSE(s)) - \log_{10}(MSE(s \setminus \{i\}))], \ s \in S_i, \quad (S.5)$$

where $s \setminus \{i\}$ is the new set produced by removing rule *i* from rule coalition *s*; and *MSE(s)* and *MSE(s\\{i})* refer to the MSE of the model trained with rule coalition *s* and $s \setminus \{i\}$, respectively. Considering that the change of MSE usually strides across orders of magnitude, the logarithm of MSE is adapted to measure the contribution. Eq. (S.5) can be simplified into:

$$\delta_i^{RI}(s) = \log_{10}(\frac{MSE(s \setminus \{i\})}{MSE(s)}), \ s \in S_i. \tag{S.6}$$

Then, the rule importance is calculated by the average of the defined marginal contribution, which can be written as:

$$RI_i = \frac{1}{N_{S_i}} \sum_{s \in S_i} \delta_i^{RI}(s), \tag{S.7}$$

where $RI_i$ is the importance of rule $i$; and $N_{S_i}$ is the size of $S_i$. For each rule $i$, there are two possible situations, namely, presence or absence, which can be represented by the binary code 1 or 0, respectively. Therefore, for $N$ considered rules, the possible coalition can be represented by a binary sequence with the length of $N$. Consequently, the $N_{S_i}$ is kept the same as $2^{N-1}$, where $N$ is the number of considered rules. Finally, the formula for calculating the rule importance ($RI$) can be obtained as follows:

$$RI_i = \frac{1}{2^{N-1}} \sum_{s \in S_i} \log_{10}(\frac{MSE(s \setminus \{i\})}{MSE(s)}). \tag{S.8}$$

For comparison, we also define the *full importance* (*FI*) in this work, which refers to the importance of a given rule when all of the other rules exist. The formula for *FI* can be expressed as:

$$FI_i = \log_{10}\left(\frac{MSE(s_f \setminus \{i\})}{MSE(s_f)}\right), \tag{S.9}$$

where $FI_i$ is the full importance of rule $i$; and $s_f$ is the rule coalition that includes all rules.

**Calculation process for rule importance**

As derived above, the formula of rule importance is written as Eq. (S.7) and Eq. (S.8). A visual example of calculating rule importance is provided in Fig. S1A where four rules are incorporated. For the PDE rule (rule 1), all coalitions that include and exclude this rule are considered. For each possible coalition, a machine learning model is trained to obtain the respective $MSE(s)$ and $MSE(s \setminus \{i=1\})$. In brief, the influence of the rule in all relevant rule coalitions is employed to calculate the marginal contribution as its importance. The whole calculation process is provided in Fig. S1B. In this example, a deep neural network model is utilized, and due to the computational cost associated with training multiple models, a disturbance-based approach is employed. First, a neural network model is trained without any rules until convergence, which serves as a baseline. Then, relevant rule coalitions are incorporated into the pre-trained model, respectively, to disturb the convergence. The model is trained for several epochs until a new convergence is established. Finally, the MSE of each informed model on the test data is measured to calculate the rule importance by Eq. (S.8). Similar to the Shapley value, the proposed rule importance can be accelerated by the Monte Carlo-type method (*27*).

From Eq. (S.8), it can be seen that calculating the importance of each rule requires training $2^N$ models, which covers all possible rule alliances. Considering the computational cost of model training, it is unacceptable to retrain each model with different rule alliances. Therefore, a more efficient method is proposed in this work to accelerate this process. As illustrated in Fig. S1B, a model is pre-trained without any rule for sufficient epochs until the training loss converges. Then, the different rule alliance is incorporated into the networks independently to disturb the existing convergence. An example is provided in Fig. S1, where the Burgers' shock equation is taken as an example. 100 data are sampled as the training data, and the involved rules include the governing partial differential equations (PDEs), initial conditions (ICs), and boundary conditions (BCs). In the pre-training phase, the training data loss continues to decrease and tends to converge; whereas, the rule loss remains elevated. At this time, the rule alliance is added to the model and disturbs the convergence (Fig. S1C). After the incorporation of the rule alliances, the data loss presents a sudden enlargement and starts to decrease together with the rule loss. Meanwhile, as shown in Fig. S1D, the validating loss achieves a convergence in the pre-train stage rapidly. After adding rules, a new convergence is achieved after certain epochs, and the convergence value is less than that in the pre-train stage. Therefore, in practice, the parameters of the pre-trained network and optimizer are saved, and the involved rule alliances are added to the pre-trained network to continue training with the constraint of rules. Compared with training a complete physical constraint model for each rule alliance, this strategy can save considerable time by experiencing the long pre-train process only once.

**Experimental settings**

Our interpretable framework of evaluating the worth of knowledge is essentially model-agnostic since it only relies on the input and output of the model. Therefore, it can be combined with a variety of common network architectures. In this work, we take the conventional fully connected neural network as an example. For the four canonical PDEs, a five-layer artificial neural network (ANN) with 50 neurons in each hidden layer is employed. The activation function is the Sin function for the KdV equation, the Burgers' shock equation, and the KG equation, and the Tanh function for the convection-diffusion equation. For the two-dimensional PDE and the multivariable equation, a three-layer artificial neural network (ANN) with 50 neurons in each hidden layer is employed. The activation function is ReLU. For all cases, the optimizer is Adam with a learning rate of 0.001. The rules are incorporated into the model through the loss function. Specifically, each rule is converted to a loss and added to the loss function. It is worth noting that the loss of rules is calculated on the collocation points, the settings of which are detailed in Supplementary Information S1.

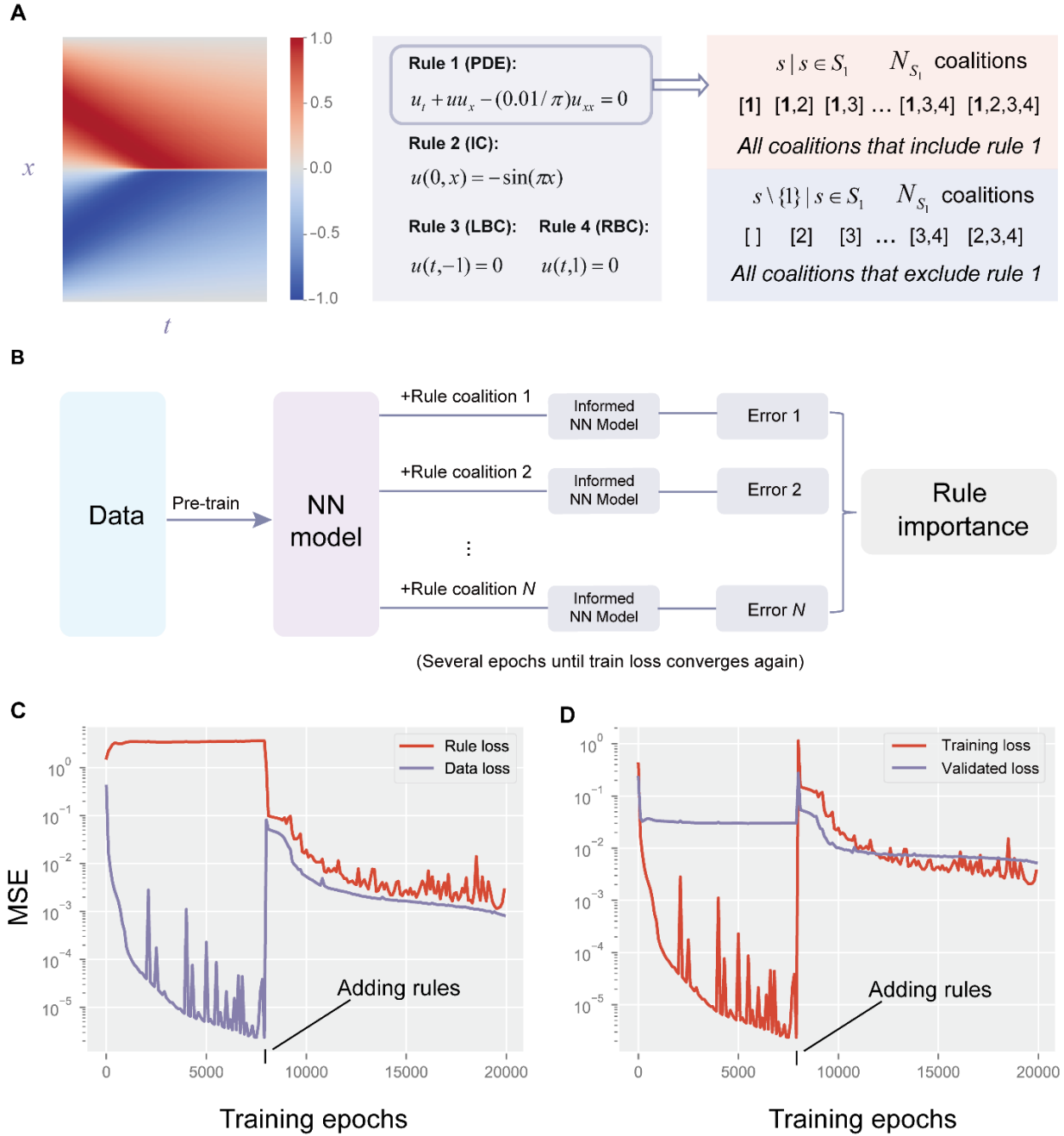

**Fig. S1. Illustration of the calculation process for rule importance.** (**A**) An example of calculating the importance of the PDE rule for Burgers' shock equation. Here, PDE refers to the partial differential equation, LBC refers to the left boundary condition, RBC refers to the right boundary condition, and IC refers to the initial condition. (**B**) The flow chart for calculating the rule importance. (**C**) The rule loss and data loss on the training data. The rules are added to the network at 8,000 epochs where convergence has been achieved in the pre-train stage. (**D**) The training loss and validating loss during the training process. The training loss only contains the data loss in the pre-train stage, and contains both data loss and rule loss after the rules are incorporated into the network. The validated loss is the mean squared error between the predicted and observed validating data.

# Supplementary Text
## 1. Supplementary information for experiments

In the manuscript, several numerical experiments have been conducted to explore the value of knowledge under various affecting factors, including data volume, data noise, in-distribution prediction, and out-distribution prediction. In this section, the experimental settings of the four canonical PDEs are detailed for the reproduction of experiments. The rules involved in the following cases are illustrated in Fig. S2.

### 1.1 The Burgers' shock equation

The form of Burgers' equation is written as:

$$u_t = -uu_x + au_{xx}, \qquad (S.10)$$

where $a$ is the coefficient of the viscous term. In this work, $a$ is set to be $\frac{0.01}{\pi}$, which means that the viscous term is so small that a shock wave will emerge. Therefore, we also denote this equation as the Burgers' shock equation. The prediction of $u$ when the shock wave emerges is difficult since $u$ changes dramatically at the shock wave. The dataset is generated by numerical simulation. For the dataset, there are 256 spatial observation points in $x \in [-1,1]$, and 100 temporal observation points in $t \in [0,1)$. Consequently, the total number of data is 25,600. In this case, the conditions of interpolation and extrapolation are both considered. For interpolation, the training data are randomly sampled from the data in the temporal domain $t \in [0,1)$, while the test data are also sampled from the remaining data in the temporal domain $t \in [0,1)$. For extrapolation, the training data are randomly sampled from the data in the temporal domain $t \in [0,0.5)$, while the test data are sampled from the remaining data in the temporal domain $t \in [0.5,1)$. It is worth noting that the rule loss is calculated on the collocation data that are generated by the network. The collocation points are generated on the $200 \times 200$ grid on the whole domain $x \in [-1,1]$ and $t \in [0,1)$. In this way, the incorporated rules can have a constraining effect on the whole domain. Moreover, the utilization of collocation data guarantees that the locations with no observation data also comply with the rules.

### 1.2 The Korteweg-de Vries (KdV) equation

The KdV equation is a partial differential equation discovered by Korteweg and de Vries to describe the motion of unidirectional shallow water, and is written as:

$$u_t = -uu_x - 0.0025u_{xxx}. \qquad (S.11)$$

The dataset is generated by numerical simulation with 512 spatial observation points in the domain $x \in [-1,1)$ and 201 temporal observation points in the domain $t \in [0,1]$. Therefore, the data volume is 102,912. For interpolation, the training data and test data are both randomly sampled from the data in the temporal domain $t \in [0,1]$. For extrapolation, the training data are randomly sampled from the data in the temporal domain $t \in [0,0.5)$, while the test data are sampled from the remaining data in the temporal domain $t \in [0.5,1]$.

### 1.3 The Klein-Gordon (KG) equation

The KG equation was first proposed by Oskar Klein and Walter Gordon in 1926 to describe the behavior of electrons in relativistic settings, and is written as:

$$u_{tt} = 0.5u_{xx} - 5u. \qquad (S.12)$$

The dataset is grid data of 201 spatial observation points in the domain $x \in [-1,1]$ and 201

temporal observation points in the domain $t \in [0,3]$, and thus the data size is 40,401. For interpolation, the training data and test data are both randomly sampled from the data in the temporal domain $t \in [0,3]$. For extrapolation, the training data are randomly sampled from the data in the temporal domain $t \in [0,1.5)$, while the test data are sampled from the remaining data in the temporal domain $t \in [1.5,3]$.

### 1.4 The convection-diffusion equation
The convection-diffusion equation can be employed to describe the transport of substances in fluid, such as the contaminant transport process. The governing equation is expressed as:
$$u_t = -u_x + 0.25 u_{xx}. \tag{S.13}$$
The dataset is grid data of 256 spatial observation points in the domain $x \in [0,2]$ and 100 temporal observation points in the domain $t \in [0,1]$, and thus the data size is 25,600. For interpolation, the training data and test data are both randomly sampled from the data in the temporal domain $t \in [0,1]$. For extrapolation, the training data are randomly sampled from the data in the temporal domain $t \in [0,0.5)$, while the test data are sampled from the remaining data in the temporal domain $t \in [0.5,3]$.

| Burgers' shock Equation | KdV Equation | KG Equation | Convection-diffusion Equation |
|---|---|---|---|
| **Rule 1 (PDE):** $u_t + u u_x - (0.01/\pi) u_{xx} = 0$ | **Rule 1 (PDE):** $u_t + u u_x + 0.0025 u_{xxx} = 0$ | **Rule 1 (PDE):** $u_{tt} + 5u - 0.5 u_{xx} = 0$ | **Rule 1 (PDE):** $u_t + u_x - 0.25 u_{xx} = 0$ |
| **Rule 2 (IC):** $u(0,x) = -\sin(\pi x)$ | **Rule 2 (IC):** $u(0,x) = \cos(\pi x)$ | **Rule 2 (IC):** $u(0,x) = e^{-20 x^2}(\sin(\pi x) + \sin(2\pi x))$ | **Rule 2 (IC):** $u(0,x) = \sin(\pi x) e^{-x}$ |
| **Rule 3 (LBC):** $u(t,-1)=0$ **Rule 4 (RBC):** $u(t,1)=0$ | **Rule 3 (LBC and RBC):** $u(t,-1) = u(t,1)$ | **Rule 3 (LBC):** $u(t,-1)=0$ **Rule 4 (RBC):** $u(t,1)=0$ **Rule 5:** $u_t(0,x)=0$ | **Rule 3 (LBC):** $u(t,0)=0$ **Rule 4 (RBC):** $u(t,2)=0$ |
| $x \in [-1,1]$ $t \in [0,1]$ | $x \in [-1,1]$ $t \in [0,1]$ | $x \in [-1,1]$ $t \in [0,3]$ | $x \in [0,2]$ $t \in [0,1]$ |

**Fig. S2. The form of involved rules in the four canonical PDEs.** The rule of PDE, the initial condition, left boundary condition, and right boundary condition are denoted in parentheses. The domain definition is provided in the green background. For the KdV equation, the periodic boundary condition is utilized, which is denoted as both LBC and RBC.

### 2. Supplementary experiments
### 2.1 The significance of collocation points
Collocation points are crucial for the function of integrated rules. The main idea of collocation points is that they are simple, yet efficient, in that the rules can restrict the sample distribution in the whole definition domain, rather than just in the observable domain. One of the advantages of the neural network is that it can provide the inference at any point, which means that the rule can not only constrain observation points, but also constrain inference points. Therefore, in practice, the collocation points are sampled on the grid points in the whole domain. In order to better demonstrate the significance of collocation points, several experiments are conducted in this section. The influence of the number of collocation points is investigated. The Burgers' shock equation in interpolation is taken as an example, and the experimental settings are kept the same as those described in Section 2.1. In the manuscript, the collocation points are the grid data of 200 × 200. Here, the rule importance with different numbers of collocation points is calculated, including 10×10, 50×50, 100×100, 200×200, and 500×500. The results are displayed in Fig. S3. It can be seen that the rule importance exhibits a general uptrend when the number of collocation points increases, especially for the rules of PDE and the initial condition. This experiment indicates that more collocation points bring a larger restriction ability of rules. However, considering that more collocation points require more computational cost in the training process, the balance between accuracy and efficiency should be carefully evaluated.

Another experiment is conducted to investigate the influence of the choice of collocation points. The example of solving multivariable equations is taken here, the setting of which is the same as that described in the manuscript. Here, the choice of collocation points is changed and compared in three situations, including interpolation with outer collocation points, interpolation with inner collocation points, and extrapolation with outer collocation points. For interpolation, the training and test data are both sampled from the definition domain of $a \in [0, \pi]$ and $b \in [-\pi, 0]$. For extrapolation, the training data are sampled from the definition domain, while the test data are sampled from $a \in [\pi, 2\pi]$ and $b \in [\pi, 2\pi]$. For inner collocation points, the sampled domain is $a \in [0, \pi]$ and $b \in [-\pi, 0]$, while the sampled domain is $a \in [0, 2\pi]$ and $b \in [-\pi, 2\pi]$ for outer collocation points. It is obvious that the outer collocation points constrain the whole domain, while the inter collocation points only constrain the interpolation domain. The corresponding rule importance is displayed in Fig. S4. It can be seen that the choice of collocation points highly affects the rule importance, which indicates that the extrapolation ability of the informed model mainly comes from the constraints of rules in the entire definition domain. If the collocation points only encounter the interpolation domain (i.e., inner collocation points), the rule importance will be sharply decreased. This experiment also exposes the drawbacks of informed machine learning, in that it highly depends on the constraints brought by the rules on the whole domain. Faced with complicated rules that cannot be converted into mathematical constraints or collocation points that cannot be sampled on the unobservable domain, the extrapolation ability of the informed model may be affected.

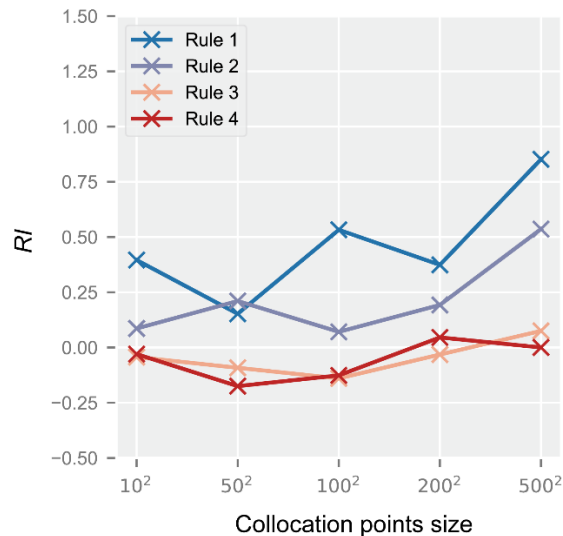

**Fig. S3. The rule importance with different collocation points size.** Here, the collocation points are sampled on the grid in the whole domain, for example, $10^2$ refers to 10×10. Rule 1 is the PDE rule, Rule 2 is the initial rule, and Rules 3 and 4 are the condition rules.

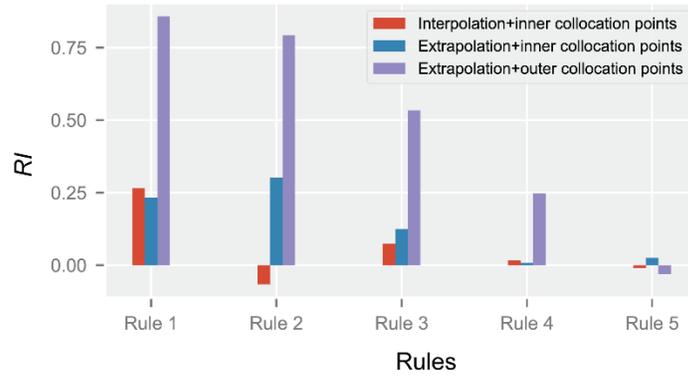

**Fig. S4. The influence of the choice of collocation points.** The example of solving multivariable equations is taken here.

**2.2 Rule importance for solving multivariable equations and 2D partial differential equations**
In this section, the rule importance of solving multivariable equations and 2D partial differential equations is calculated and provided. For the multivariable equations, we consider four conditions, including no data, 100 data under interpolation, and extrapolation, respectively. For interpolation, the training and test data are both sampled from the definition domain of $a$ and $b$. For extrapolation, the training data are sampled from the definition domain, while the test data are sampled from $a \in [\pi, 2\pi]$ and $b \in [\pi, 2\pi]$. The results are shown in Fig. S5A. It can be seen that the implicit rule (rule 5) has the least importance, which indicates that the implicit range limit has been learned well through the data and other rules. Therefore, rule 5 is highly replaceable. Meanwhile, in this case, the testing domain of extrapolation is far away from the training data, which implies that the sample distributions differ significantly. Therefore, the rule importance under extrapolation is generally lower than that under interpolation with no data. This inference can be confirmed by the fact that adding 100 data assists to improve the rule importance under extrapolation. This result reflects the limitation of mere rules since their generalization ability is restricted without data. For interpolation, the displayed principle is parallel with that of the previous section.

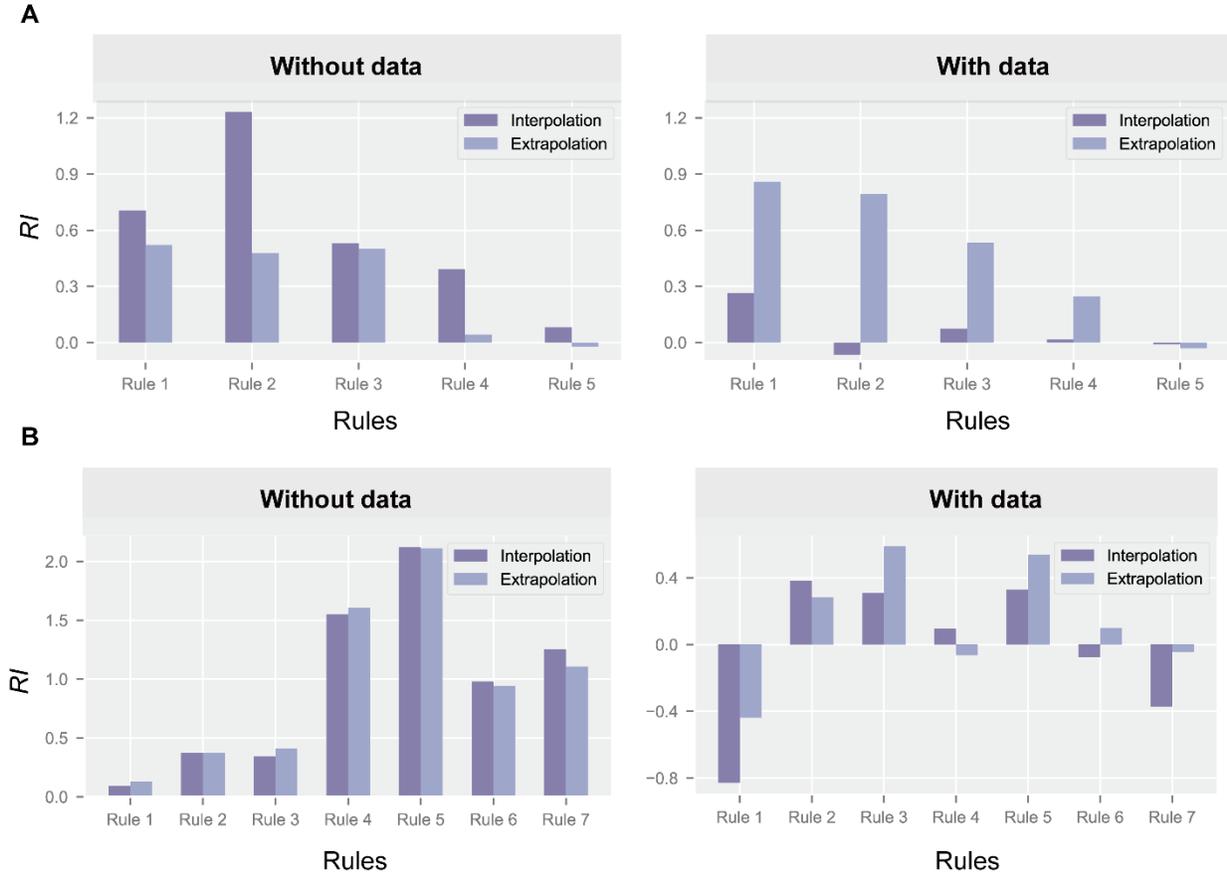

**Fig. S5. Rule importance of solving multivariable equations and two-dimensional partial differential equations (PDEs).** **(A)** The importance of the involved rules in the cases of interpolation and extrapolation with and without data when solving multivariable equations, respectively. **(B)** The importance of the involved rules in the cases of interpolation and extrapolation with and without data when solving the two-dimensional partial differential equations (PDEs), respectively.

For solving 2D partial differential equations, the interpolation and extrapolation situations are both considered for comparison. For interpolation, the training data and test data are both sampled from the whole definition domain. For extrapolation, the test data are sampled from $x \in [0.25, 0.75]$ and $y \in [0.25, 0.75]$, while the training data are sampled from the remaining space in the definition domain. Rule importance in the case of interpolation and extrapolation with no data and 50 data is depicted in Fig. S5B, respectively. It is confirmed that the overall principle of this case is parallel with that obtained from the previous section. Moreover, it can be seen that the rule importance is almost not affected by extrapolation or interpolation when there are no data. Under this circumstance, the sample distribution can only be learned by rules without data, which equals solving the PDE under the given conditions.

**2.3 Rule importance for the practical 2D heterogeneous subsurface flow problem**
In this section, the function of rule importance in practical complex problems is verified by a specific case in geoscience: heterogeneous subsurface flow. The case is obtained from Wang et al. (*28*), in which a theory-guided neural network is invented to handle this complex engineering problem. As illustrated in Fig. S6, the governing equation is written as:

$$S_s \frac{\partial h}{\partial t} = K \frac{\partial^2 h}{\partial x^2} + K \frac{\partial^2 h}{\partial y^2}, \quad (S.14)$$

where $S_s$ denotes specific storage; $K$ denotes the heterogeneous hydraulic conductivity; and $h$ denotes the hydraulic head. The hydraulic conductivity $K$ is a random field that varies according to the spatial domain and thus can be denoted as $K(x,y)$, which is illustrated in Fig. S6A. The experimental setting is the same as that in Wang et al. (28). As a complex engineering case, several rules can be employed to facilitate the training of the model, including the governing equation (rule 1), boundary equation (rule 2), expert knowledge (rules 3 and 4), initial condition (rule 5), and engineering control (rule 6). The form of the rules is displayed in Fig. S6A. The coefficients of rules are set to be 1 by default. In the experiment, the first 10 time-steps of hydraulic head $h$ are known as the training data, and the subsequent 40 time-steps are treated as the test data. The utilized neural network is an ANN with seven layers, where each hidden layer has 50 neurons. The optimizer is Adam with 2,000 training epochs. The collocation points are sampled from the whole domain (0 to 50 time-steps). The importance of each rule is calculated and illustrated in Fig. S6B. From the figure, it is evident that the boundary condition presents the highest rule importance. Meanwhile, the initial condition shows negative importance. This may be caused by the fact that the information on the initial condition has been contained in the training data since the initial hydraulic head $h$ is known. Moreover, the governing equation and the boundary condition mainly determine the future development of the hydraulic head $h$. Regarding the governing equation rule, the influence of the number of the relying rules is displayed in Fig. S6C. It can be seen that the PDE rule needs other rules to produce its effect since it requires at least four other rules to generate high rule importance. This experiment confirms that the proposed rule importance can handle complicated practical scenes well to reveal the worth of knowledge.

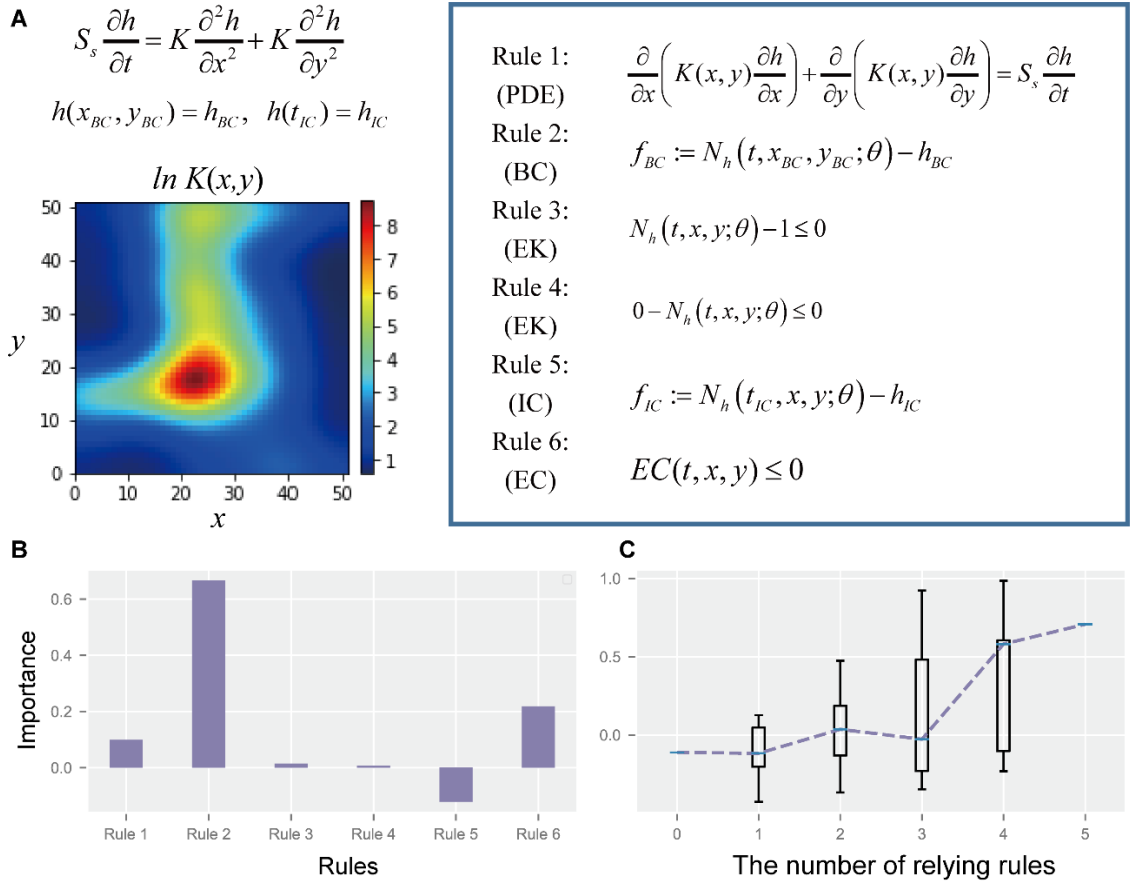

**Fig. S6. Worth of knowledge in a practical complicated engineering scene.** (**A**) The governing equation of the heterogeneous subsurface flow, the visualization of heterogeneous hydraulic conductivity $K$, and the formula of the involved rules. (**B**) The importance of six rules in the theory-guided neural network when predicting the future hydraulic head. (**C**) The importance of the PDE rule with different numbers of relying rules.

### 2.4 The influence of data noise on the worth of knowledge

In this section, the influence of data noise on rule importance is considered. Here, the data volume is fixed at 100, and the Gaussian noise is added to the observation data in the form denoted as:

$$\tilde{u} = u + \varepsilon \cdot std(u) \cdot N(0,1), \tag{S.15}$$

where $\tilde{u}$ are the noisy data; $u$ are the clean data; $\varepsilon$ is the noise level; and $N(0,1)$ is the normal distribution. Different levels of data noise are exerted, including 0 (clean data), 0.1 (10%), 0.2 (20%), 0.3 (30%), 0.4 (40%), and 0.5 (50%).

For the four canonical physical processes under in-distribution and out-distribution prediction, the importance of each integrated rule is calculated and illustrated in Fig. S7B and Fig. S8B. Among all types of rules, the PDE rules present the most apparent pattern, which generally follows an increasing trend when faced with moderate data noise. Specifically, a conspicuous increase emerges as the noise level rises from 0 to 0.3, while a slight descent occurs when the noise level is too large. It is interesting to discover that the rule importance of these physical processes becomes closer as the noise level increases. Meanwhile, it is found that the highest importance still occurs with moderate data noise. This phenomenon can be elucidated from the aspect of the sample distribution. In essence, data noise will influence the model's generalization ability because of the deviation in the training data. However, moderate noise may weaken the influence of overfitting (*29*). Under this circumstance, integrated rules can assist to redress the biased learned sample distribution from noisy data to some extent since they restrict the optimization space. Furthermore, it is also discovered that the PDE importance is generally higher than that of in-distribution prediction, while the robustness to data noise is influenced since the point with the highest rule importance comes earlier. This phenomenon also appears in the initial and boundary conditions (Fig. S8B). This may be attributable to the difficulty of the out-distribution prediction.

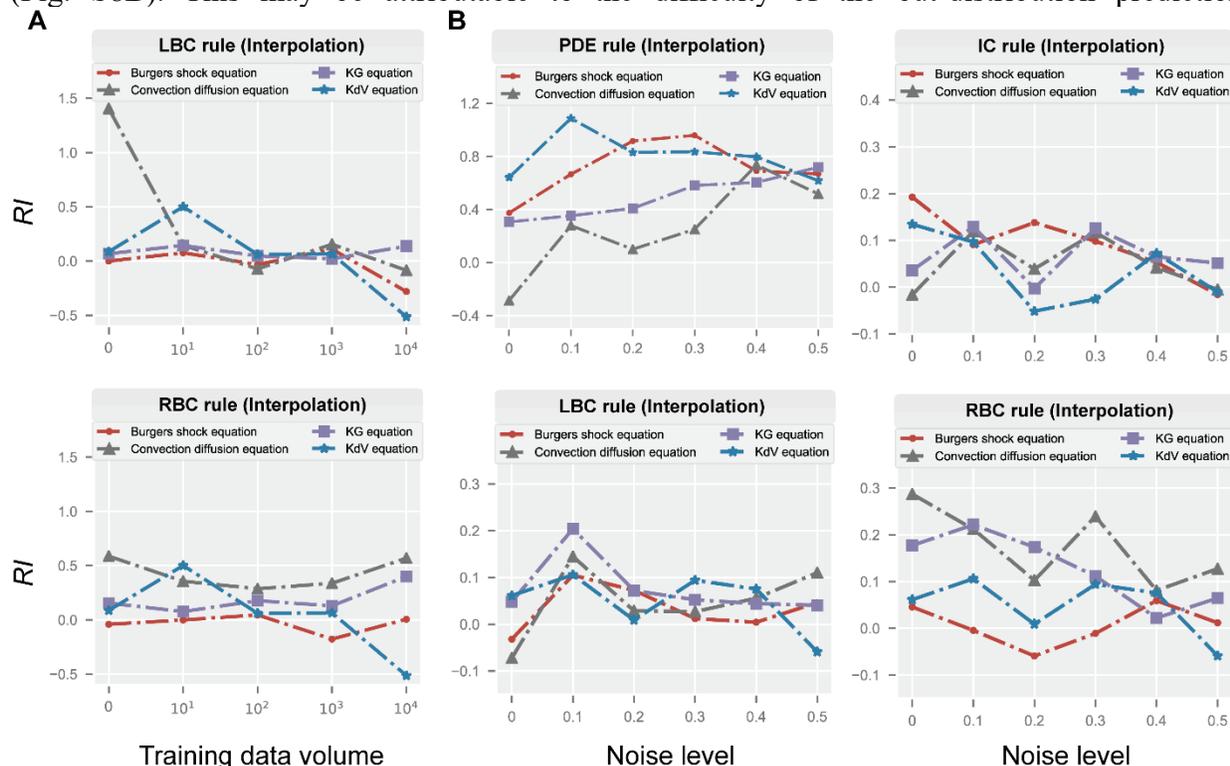

**Fig. S7. The influence of training data volume and data noise on rule importance under in-distribution (interpolation) prediction.** (**A**) The rule importance varies with the data volume in the four physical processes. (**B**) The rule importance varies with the data noise in the four physical processes. Here, LBC refers to the left boundary condition, RBC refers to the right boundary condition, IC refers to the initial condition, and PDE refers to the partial differential equation.

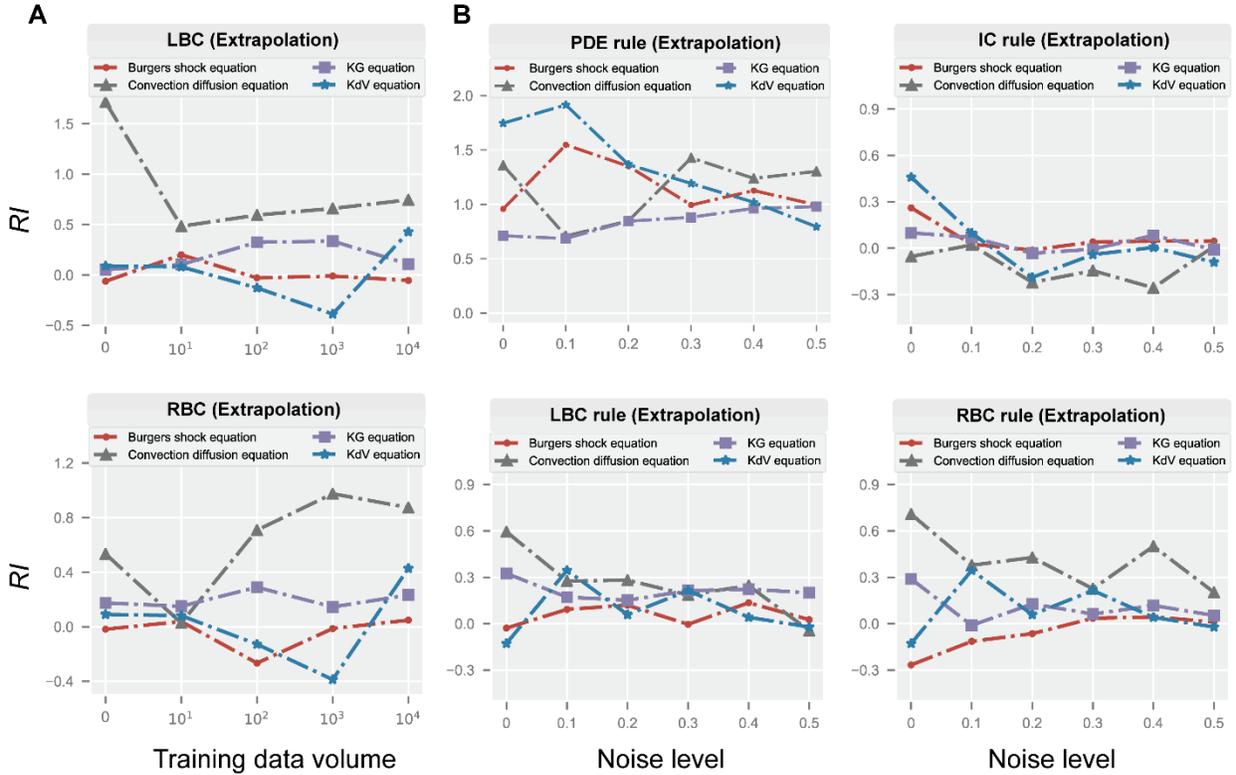

**Fig. S8. The influence of training data volume and data noise on rule importance under out-distribution (extrapolation) prediction.** (**A**) The rule importance varies with the data volume in the four physical processes. (**B**) The rule importance varies with the data noise in the four physical processes. Here, LBC refers to the left boundary condition, RBC refers to the right boundary condition, IC refers to the initial condition, and PDE refers to the partial differential equation.

## 3. Supplementary figures and tables

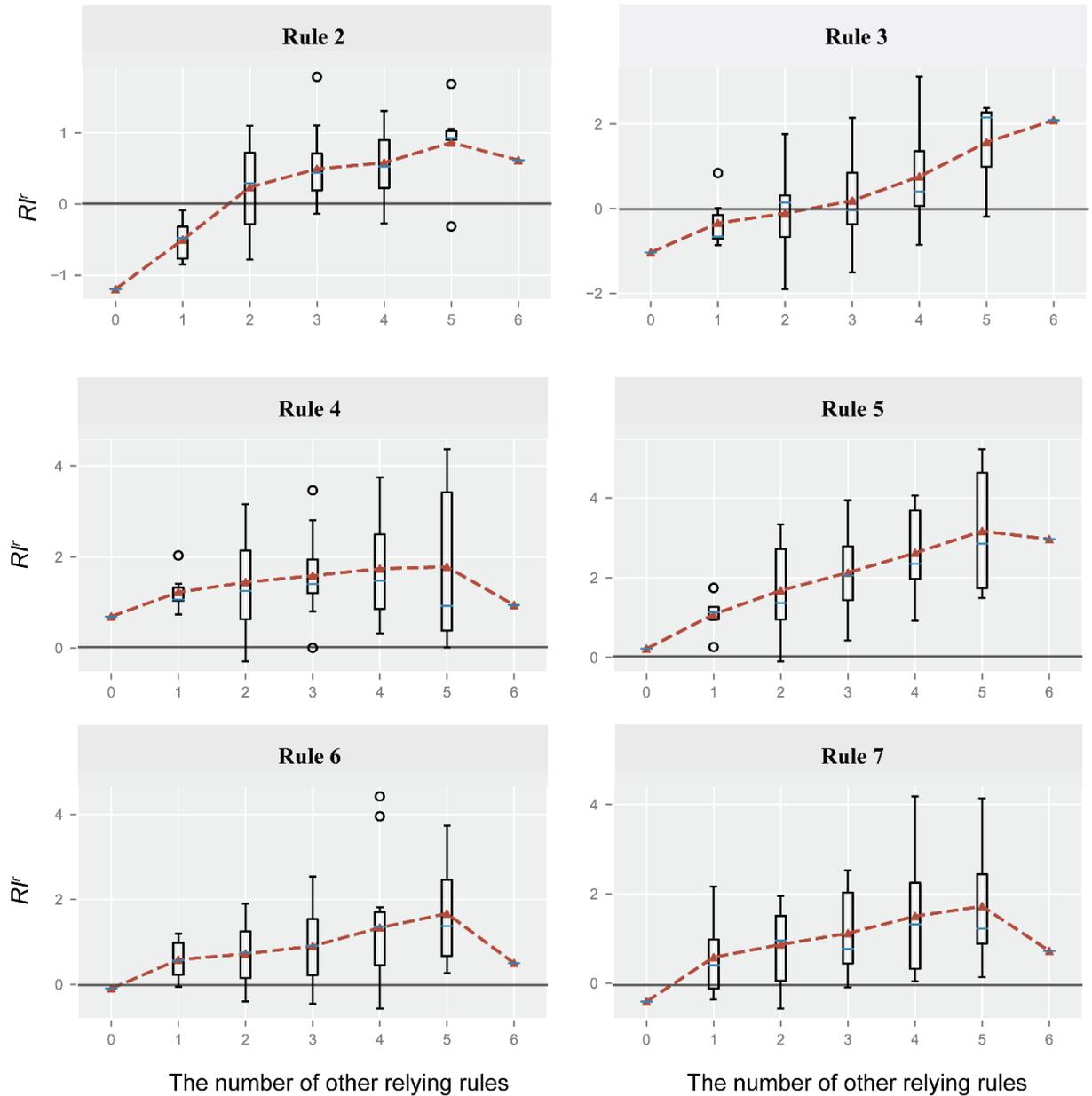

**Fig. S9. The importance of the rules under different numbers of relying rules without data when solving two-dimensional PDEs.** The definition of rule 2 to rule 7 is demonstrated in Eq. (7) in the manuscript. The dark solid line refers to zero importance.

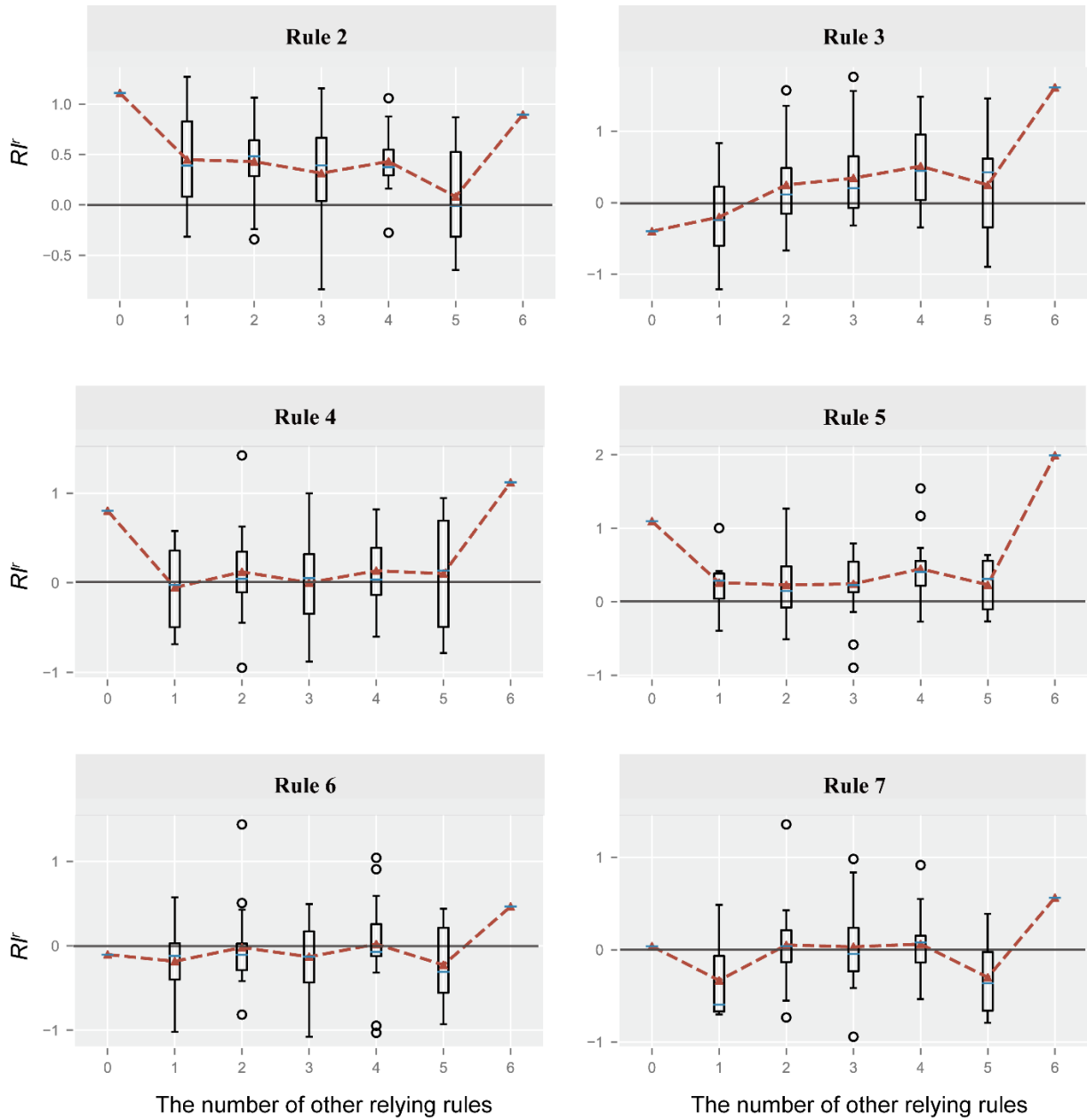

**Fig. S10. The importance of the rules under different numbers of relying rules with data when solving two-dimensional PDEs.** The definition of rule 2 to rule 7 is demonstrated in Eq. (7) in the manuscript. The dark solid line refers to zero importance.

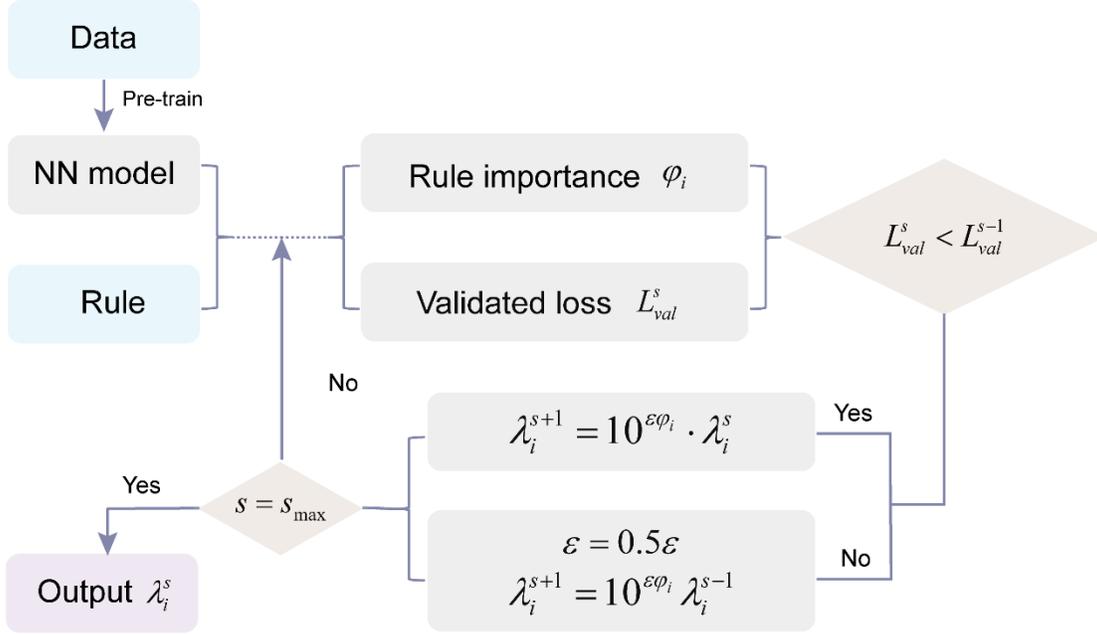

**Fig. S11. The flow chart for optimizing the weights of rules.** Here, $s$ is the iteration, $i$ is the rule index, and $\lambda_i^s$ refers to the weight of rule $i$ in the iteration $s$. $\varepsilon$ is the optimization step size, which is diminished when the validating loss does not decrease.

**Table S1. The comparison between different ways of adjusting weights $\lambda_i$ of rule $i$.** The optimal weights found by different methods are displayed in the table. The test MSE is the mean squared error on the test data. Gradient flow is an adaptive method (7), and the $\lambda_i$ is not fixed. The empirical method adjusts the loss of each item to the same order of magnitude. The bold column refers to the best method with the smallest testing MSE.

| Optimize method | Default | Gradient flow | Empirical | Ours |
|---|---|---|---|---|
| $\lambda_1$ | 1 | -- | 1 | **2.11** |
| $\lambda_2$ | 1 | -- | 1 | **1.02** |
| $\lambda_3$ | 1 | -- | 100 | **1.26** |
| $\lambda_4$ | 1 | -- | 1 | **1.18** |
| $\lambda_5$ | 1 | -- | 10000 | **1.02** |
| **Test MSE** | $1.9\times10^{-4}$ | $6.0\times10^{-5}$ | $7.0\times10^{-4}$ | **$4.9\times10^{-5}$** |